\documentclass[11pt]{article}
\usepackage{coling2020}
\usepackage{times}
\usepackage{url}
\usepackage{hyperref}
\usepackage{latexsym}
\usepackage{multirow}
\usepackage{amssymb}
\usepackage{amsmath}
\usepackage[pdftex]{graphicx}
\usepackage{subcaption}
\usepackage{microtype}

\colingfinalcopy 

\usepackage{array}
\newcolumntype{P}[1]{>{\centering\arraybackslash}p{#1}}

\title{Neural Machine Translation Doesn't Translate Gender Coreference Right Unless You Make It}
\author{Danielle Saunders \and Rosie Sallis \and Bill Byrne \\
    Department of Engineering, University of Cambridge, UK  \\
      {\tt \{ds636, rs965, wjb31\}@cam.ac.uk}}

\begin{document}

\maketitle
\begin{abstract}
Neural Machine Translation (NMT) has been shown to struggle with grammatical gender that is dependent on the gender of human referents, which can cause gender bias effects. Many existing approaches to this problem seek to control gender inflection in the target language by explicitly or implicitly adding a gender feature to the source sentence, usually at the sentence level.  

In this paper we propose schemes for incorporating explicit word-level gender inflection tags into NMT. We explore the potential of this gender-inflection controlled translation when the gender feature can be determined from a human reference, or when a test sentence can be automatically gender-tagged, assessing on English-to-Spanish and English-to-German translation.

We find that simple existing approaches can over-generalize a gender-feature to multiple entities in a sentence, and suggest effective alternatives in the form of tagged coreference adaptation data. We also propose an extension to assess translations of gender-neutral entities from English given a corresponding linguistic convention, such as a non-binary inflection, in the target language.

\end{abstract}

\section{Introduction}

%
%
\blfootnote{
    %
    %
    %
    %
    \hspace{-0.65cm}  
    This work is licensed under a Creative Commons 
    Attribution 4.0 International Licence.
    Licence details:
    \url{http://creativecommons.org/licenses/by/4.0/}.
    
    %
}

Translation into languages with grammatical gender involves correctly inferring the grammatical gender of all entities in a sentence. In some languages this grammatical gender is dependent on the social gender of human referents. For example, in the Spanish translation of the sentence `This is the doctor',  `the doctor' would be either  `el médico', masculine, or `la médica', feminine. Since the noun refers to a person the grammatical gender inflection should be correct for a given referent. 

In practice many NMT models struggle at generating such inflections correctly \cite{sun-etal-2019-mitigating}, often instead defaulting to gender-based social stereotypes \cite{prates2019assessing} or masculine language \cite{hovy-etal-2020-sound}. For example, an NMT model might always translate `This is the doctor' into a sentence with a masculine inflected noun: `Este es el médico'.

Such behaviour can be viewed as translations  exhibiting gender bias. By `bias' we follow the definition from \newcite{friedmanbias1996} of behaviour which `systematically and unfairly  discriminate[s]  against certain individuals or groups of individuals in favor of others.' Specifically, translation performance favors referents fitting into groups corresponding to social stereotypes, such as male doctors. 

Such systems propagate the representational harm of erasure to referents -- for example, a non-male doctor would be incorrectly gendered by the above example translation. Systems may also cause allocational harms if the incorrect translations are used as inputs to other systems \cite{crawford2017trouble}. System users also experience representational harms via the reinforcement of stereotypes associating occupations with a particular gender \cite{abbasi2019fairness}. Even if they are not the referent, the user may not wish for their words to be translated in such a way that they  appear to endorse social stereotypes. Users will also experience a lower quality of service in receiving grammatically incorrect translations. 

A common approach to this broad problem in NMT is the use of gender features, implicit or explicit. The gender of one or more words in a test sentence  is determined from external context \cite{vanmassenhove-etal-2018-getting,basta-etal-2020-towards} or by reliance on `gender signals' from words in the source sentence such as gendered pronouns. That information can then be used when translating. Such approaches combine two distinct tasks: identifying the gender inflection feature, and then applying it to translate words in the source sentence. These feature-based approaches make the unstated assumption that if we \emph{could} correctly identify that, e.g., the doctor in the above example should be female, we could inflect entities in the sentence correctly, reducing the effect of gender bias. 

Our contribution is an exploration of this assumption. We propose a scheme for incorporating an explicit gender inflection tag into NMT, particularly for translating coreference sentences \emph{where the reference gender label is known}. Experimenting with translation from English to Spanish and English to German, we find that simple existing approaches overgeneralize from a gender signal, incorrectly using the same inflection for every entity in the sentence. We show that a tagged-coreference adaptation approach is effective for combatting this behaviour.  Although we only work with English source sentences to extend prior work, we note that our approach can be extended to source languages without inherent gender signals like gendered pronouns, unlike approaches that rely on those signals.

Intuitively, if gender tagging does not perform well when it can use the label determined by human coreference resolution, it will be even less useful when a gender label must be automatically inferred.  Conversely, gender tagging that is effective in this scenario may be beneficial when the user can specify the gendered language to use for the referent, such as Google Translate's translation inflection selection \cite{johnson2018providing}, or for translations where the grammatical gender to use for  all human referents is known.  We also find that our approach works well  with RoBERTa-based gender tagging for English test sentences.

Existing work in NMT gender bias has focused on the translation of sentences based on binary gender signals, such as exclusively male or female personal pronouns. This excludes and erases those who do not use binary gendered language, including but not limited to non-binary individuals \cite{zimman2017transgender,cao-daume-iii-2020-toward}. As part of this work we therefore explore applying tagging to indicate gender-neutral referents, and produce a WinoMT set to assess translation of coreference sentences with gender-neutral entities.

\subsection{Related work}
Variations on a gender tag or signal for machine translation have been proposed in several forms. \newcite{vanmassenhove-etal-2018-getting} incorporate a `speaker gender' tag into training data, allowing gender to be conveyed at the sentence level. However, this does not allow more fine-grained control, for example if there is more than one referent in a sentence. Similar approaches from \newcite{voita-etal-2018-context}  and \newcite{basta-etal-2020-towards}  infer and use gender information from discourse context. \newcite{moryossef-etal-2019-filling} also incorporate a single explicit gender feature for each sentence at inference.  \newcite{miculicich-werlen-popescu-belis-2017-using}  integrate coreference links into machine translation reranking to improve pronoun translation with cross-sentence context. \newcite{stanovsky-etal-2019-evaluating} propose NMT gender bias reduction by `mixing signals' with the addition of pro-stereotypical adjectives. Also related to our work is the very recent approach of \newcite{bergmanis2020mitigating}, who train their NMT models from scratch with all source language words annotated with target language grammatical gender.

In \newcite{saunders-byrne-2020-reducing} we treat gender bias as a domain adaptation problem by adapting to a small set of synthetic sentences with equal numbers of entities using masculine and feminine inflections. We also interpret this as a gender `tagging' approach, since the gendered terms in the synthetic dataset give a strong signal to the model. In this work we extend the synthetic datasets from this work to explore this effect further.

Other approaches to reducing gender bias effects involve adjusting the word embeddings either directly \cite{escude-font-costa-jussa-2019-equalizing} or by training with counterfactual data augmentation  \cite{zhao-etal-2018-gender,zmigrod-etal-2019-counterfactual}. We view these approaches as orthogonal to our proposed scheme: they have similar goals but do not directly control inference-time gender inflection at the word or sentence level.

\section{Assessing and controlling gender inflection}
We wish to investigate whether a system can translate into inflected languages correctly correctly given the reference gender label of a certain word. Our proposed approach involves fine-tuning a model on a very small, easily-constructed synthetic set of sentences which have  gender tags. At test time we assign the reference gender label to the words whose gender inflection we wish to control.

\subsection{Gender bias assessment}
WinoMT \cite{stanovsky-etal-2019-evaluating} is a test set for assessing the presence of gender bias in translation from English to several gender-inflected languages. Each of 3888 test sentence contains two human entities, one of which is coreferent with a pronoun. 1826 of these sentences have male primary entities, 1822 female and 240 neutral. The first test sentence in WinoMT is:

\emph{The developer argued with the designer because she did not like the design.}

The gender label for this sentence is `female' and the primary entity label is `the developer'. The same sentence with a gender tag would be:

\emph{The developer \texttt{$<$F$>$} argued with the designer because she did not like the design.}

We only  tag the primary entity in test sentences. During evaluation WinoMT extracts the hypothesis translation for `the developer' by automatic word alignment and assesses its gender inflection in the target language. The main objective is high overall accuracy -- the percentage of correctly inflected primary entities.  

We  note a comment by \newcite{rudinger-etal-2018-gender}, who develop a portion of the English WinoMT source sentences, that such schemas `may demonstrate the presence of gender bias in a system, but not prove its absence.' In fact high WinoMT accuracy can be achieved by using the labeled inflection for \emph{both} entities in a WinoMT test sentence, even though only one is specified by the sentence. 

We therefore produce\footnote{Our new adaptation and evaluation sets can be found at \url{https://github.com/DCSaunders/tagged-gender-coref}} a test set for the WinoMT framework to track the gender inflection of the secondary entity in each original WinoMT sentence (e.g. `the designer' in the above example). We measure second-entity inflection correspondence with the gender label, which we refer to as  \textbf{L2}. High L2 suggests that `the designer' would also have feminine inflection in a translation of the above example, despite not being coreferent with the pronoun.

We are particularly interested in cases where L2 increases over a baseline, or high $\mathbf{\Delta}$\textbf{L2}.  Many factors may contribute to a baseline system's L2, but we are specifically interested in whether \emph{adding} gender features affects only the words they are intended to affect. High $\Delta$L2 indicates a system learning to over-generalize from available gender features. We consider this as erasing the secondary referents, and therefore as undesirable behaviour.

\subsection{Adaptation to gender-feature datasets}
\begin{table*}[ht]
    \centering
        \small 

    \begin{tabular}{|p{0.7cm}|p{4.3cm}|p{4.2cm}|p{4.5cm}|}
    \hline
     \textbf{Name}  & \textbf{English source} & \textbf{German target} & \textbf{Spanish target}\\
     \hline
          \multirow{3}{*}{\textbf{S\&B}} & the trainer finished his work&  der Trainer  beendete seine Arbeit & el entrenador  terminó su trabajo \\
        &the trainer finished her work&  die Trainerin  beendete ihre Arbeit & la entrenadora  terminó su trabajo \\
        & the trainer finished their work &  \texttt{DEF}  Trainer\texttt{W\_END}  beendete \texttt{PRP}  Arbeit & \texttt{DEF}  entrenador\texttt{W\_END}  terminó su trabajo  \\
        \hline
        \textbf{V1} & the trainer  \texttt{$<$M$>$}  finished his work & der Trainer  beendete seine Arbeit & el entrenador  terminó su trabajo\\
        \hline
         \textbf{V2} & the trainer \texttt{$<$F$>$} finished the work & die Trainerin  beendete die Arbeit & la entrenadora  terminó el trabajo  \\
        \hline
         \textbf{V3} & the trainer \texttt{$<$N$>$} and the choreographer \texttt{$<$M$>$} finished the work &   \texttt{DEF} Trainer\texttt{W\_END} und der Choreograf  beendeten die Arbeit &  \texttt{DEF} entrenador\texttt{W\_END} y el coreógrafo terminaron el trabajo\\
        \hline
         \textbf{V4} & the trainer \texttt{$<$F$>$}, the choreographer \texttt{$<$N$>$} &  die Trainerin, \texttt{DEF} Choreograf\texttt{W\_END}& la entrenadora,  \texttt{DEF} coreógraf\texttt{W\_END}\\ 
         \hline
    \end{tabular}
    \caption{Examples of the tagging schemes explored in this paper. Adjective-based sentences (e.g. `the tall woman finished her work') are never tagged. For neutral target sentences, we define synthetic placeholder articles \texttt{DEF} and  noun inflections \texttt{W\_END}, as well as a placeholder possessive pronoun for German \texttt{PRP}}
    \label{tab:examples}
\end{table*}

\label{sec:tags}
In \newcite{saunders-byrne-2020-reducing} we propose reducing gender bias effects quickly by model adaptation to sets of 388 simple synthetic sentences with equal numbers of male and female entities. A gendered-alternative-lattice rescoring scheme avoids catastrophic forgetting. The sentences follow a template:
\begin{center}\textit{The [entity] finished [his$|$her] work.}\end{center}
In one set the \emph{entity} is always a profession (e.g. `doctor'). In the other it is either `\emph{[adjective]  [man$|$woman]}'(e.g. `tall man') or a profession that does not occur in WinoMT source sentences (e.g. `trainer'.) We use the latter set to minimize the  confounding effects of vocabulary memorization.

It is possible to extract natural text with gendered entities, for example using GeBioToolkit \cite{costa2020gebiotoolkit}. The synthetic dataset is more suited to our work for two reasons: it has been shown to allow strong accuracy improvements on WinoMT, and it has a predictable format that can easily be augmented with gender tags. We leave the more complicated scenario of extracting and tagging natural adaptation data to future work.

As well as the unchanged \textbf{S\&B} synthetic adaptation set, we propose four gender-tagged variations, which we illustrate in Table \ref{tab:examples}. In the first, \textbf{V1}, we add a gender tag following professions only (we do not tag adjective-based sentences since `man' and `woman' are already distinct words in English).

For the second, \textbf{V2}, we use the same tagging scheme but note that the possessive pronoun offers a gender signal that may conflate with the tag, so change all examples to `... finished \emph{the} work'.

The third, \textbf{V3}, is the same as \textbf{V2} but in each profession-based sentence a second profession-based entity with a different gender inflection tag is added. This is intended to discourage systems from over-generalizing one tag to all sentence entities. 

In the final scheme, \textbf{V4}, we simplify \textbf{V3} to a minimal, lexicon-like pattern: 
\begin{center}\textit{The [entity1], the [entity2].}\end{center}
Both entities are tagged. We remove all adjective-based sentences, leaving only tagged coreference profession entities for adaptation. This set has the advantage of using simpler language than other sets, making it easier to extend to new target languages.

\subsection{Exploring gender-neutral translation}
We wish to extend previous machine translation coreference research to the translation of gender-neutral language, which may be used by non-binary individuals or to avoid the social impact of using gendered language \cite{zimman2017transgender,misersky2019grammatical}. Recently \newcite{cao-daume-iii-2020-toward} have encouraged inclusion of non-binary referents in NLP coreference work. Their study focuses heavily on English, which has minimal gender inflection and where gender-neutral language such as singular \emph{they} is in increasingly common use \cite{bradley2019singular}; the authors acknowledge that `some extensions ... to languages with grammatical gender are non-trivial'. 

In particular, existing NMT gender bias test sets typically analyse behaviour in languages with grammatical gender that corresponds to a referent's gender. Translation into  these languages effectively highlights differences in translation between masculine and feminine referents, but these languages also often lack widely-accepted conventions for gender-neutral language \cite{ackerman2019syntactic,hord2016bucking}.  In some languages with binary grammatical gender it is  possible to avoid gendering referents by using passive or reflexive grammar, but such constructions can themselves invalidate individual identities \cite{auxlandtodes}.

We therefore explore a proof-of-concept scheme for translating tagged neutral language into inflected languages by introducing synthetic gender-neutral placeholder articles and noun inflections in the target language. For example, we represent the gender-neutral inflection of `el entrenador' (the trainer) as `\texttt{DEF} entrenador\texttt{W\_END}' 

A variety of gender-neutral inflections have been proposed for various grammatically gendered languages, such as \emph{e} or \emph{x} Spanish \cite{papadopoulos2019innovaciones} and Portuguese \cite{auxlandtodes} noun inflections instead of masculine \emph{o} and feminine \emph{a}. These language-specific approaches may develop in various forms across social groups and networks, and can shift over time \cite{shroy2016innovations}. Our intent is not to prescribe which should be used, but to explore an approach which in principle could be extended to various real inflection schemes.

We construct additional `neutral-augmented' versions of the adaptation sets described in \ref{sec:tags}, adding `\emph{The [adjective]  person finished [their$|$the] work}' sentences to the adjective-based sets and sentences like `\emph{The trainer \texttt{$<$N$>$} finished [their$|$the] work}' to the profession-based sets, with synthetic placeholder articles \texttt{DEF} and  inflections \texttt{W\_END} on the target side of profession sentences. We give examples for Spanish and German in Table \ref{tab:examples}.  We also construct a neutral-label-only version of WinoMT containing the 1826 unique binary templates filled with they/them/their. We report results adapting to the original and neutral-augmented sets separately for ease of comparison with prior work.

\section{Experiments}
\label{ss:data}
We use baseline Transformer models, BPE vocabularies, synthetic datasets and baseline rescoring gendered-alternative lattices  from \newcite{saunders-byrne-2020-reducing}\footnote{\url{https://github.com/DCSaunders/gender-debias}} and  follow the same adaptation scheme,   assessing on English-to-German and English-to-Spanish translation. We define gender tags as unique vocabulary items which only appear in the source sentence. We adapt to synthetic data with minibatches of 256 tokens for 64 training updates, which we found gave good results when fine-tuning on the S\&B datasets. The V3 sets have about 30\% more tokens, the V4 sets about 30\% fewer and the neutral-augmented sets about 50\% more: we adjust the adaptation steps accordingly for these cases.

For all results we rescore the baseline system gendered-alternative lattices with the listed model. This constrains the output hypothesis to be a gender-inflected version of the original baseline hypothesis. Lattice rescoring allows minimal degradation in BLEU while letting gender inflections in the hypothesis translation be varied for potentially large WinoMT accuracy increases. For the gender-neutral experiments we add synthetic inflections and articles to the lattices.

When assessing automatic test set tagging we use the RoBERTa \cite{liu2019roberta}  pronoun disambiguation function tuned on Winograd Schema Challenge data as described in Fairseq documentation\footnote{\url{https://github.com/pytorch/fairseq/tree/master/examples/roberta/wsc}}.

We wish to improve coreference without loss of general translation quality, and so assess BLEU on a separate, untagged general test set. For ease of comparison with previous work, we report general translation quality on the test sets from WMT18 (en-de) and WMT13 (en-es), reporting cased BLEU using SacreBLEU\footnote{BLEU+case.mixed+numrefs.1+smooth.exp+tok.13a+v.1.4.8} \cite{post-2018-call}.

\subsection{Measured improvements in gender accuracy are often accompanied by over-generalization}

Table \ref{tab:mfresults} gives BLEU score and primary-entity accuracy for the original, binary versions of  synthetic adaptation sets described in section \ref{sec:tags}. WinoMT test sentences have primary entities tagged with their gender label if the adaptation set had tags, and are unlabeled otherwise. We note that lattice rescoring keeps the general test set score within 0.3 BLEU of the baseline for all adaptation sets, and focus  on the variation in WinoMT performance.  

Primary-entity accuracy increases significantly over the baseline for all adaptation schemes. V3 and V4, which contain coreference examples, are most effective for en-es, while V2, which contains a single entity, is slightly more effective for en-de. This may reflect the difference in baseline quality: the stronger en-de baseline is more likely to have already seen multiple-entity sentences.

We also report $\Delta$L2, the change in the secondary entity's label correspondence compared to the baseline. High $\Delta$L2 implies that the model is over-generalizing a gender signal intended for the primary entity to the secondary entity. In other words, the gender signal intended for the primary entity has a very strong influence on the translation of the secondary entity. $\Delta$L2 does indeed increase strongly from the baseline for the S\&B and V1 systems, confirming our suspicion that these models trained on sentences with a single entity simply learn to apply any gender feature to both entities in the test sentences indiscriminately. 

Remarkably, for adaptation to S\&B and V1 datasets we found that the secondary entity is inflected to correspond with the pronoun more often than the primary entity which is labeled as coreferent with it. A possible explanation is that the secondary entity occurs at the start of the sentence in about two thirds of test sentences, compared to about one third for the primary entity. Adapting to single-entity test sets may encourage the model to simply inflect the first entity in the sentence using the gender signal.

For V2, where the source possessive pronoun is removed and the tag is the only gender signal,   $\Delta$L2 still increases significantly, although less than for V1. This indicates that even if the only signal is a  gender tag applied directly to the correct word, it may be wrongly taken as a signal to inflect other words. The V3 scheme is the most promising, with a 17\% increase in accuracy for en-de and a 30\% increase for en-es corresponding to very small changes in L2, suggesting this model minimizes over-generalization from gender features beyond the tagged word. V4 performs similarly to V3 for en-de but suffers from an L2 increase for en-es. It is possible that a lexicon-style set with tags in every example may cause undesirable over-generalisation.

\subsection{Reference labeled, auto-labeled and unlabeled test sentences}
\begin{table*}
    \centering
    \small 

    \begin{tabular}{|c|c|ccc|ccc|}
    \hline
         \textbf{System}  & \textbf{Labeled WinoMT} &  \multicolumn{3}{c|}{\textbf{en-de}} &  \multicolumn{3}{c|}{\textbf{en-es}}\\
         \hline
         & & BLEU &  Acc &  $\Delta$L2 & BLEU &  Acc &   $\Delta$L2\\
         Baseline   & $\times$ & 42.7 & 60.1 & - & 27.8 & 49.6& - \\
         S\&B  & $\times$ & 42.4 & 82.3 & 27.4 & 27.7 & 66.3 & 29.7 \\
         V1  & \checkmark & 42.5  &81.7 & 26.6  & 27.7 &69.0 & 26.4\\
         V2  & \checkmark& 42.5  &\textbf{84.1} & 24.2 & 27.5 & 70.9 & 13.2 \\
        V3 & \checkmark & 42.6 & 77.4  & \textbf{1.1}& 27.5 &  80.6 & \textbf{0.3} \\
        V4 & \checkmark & 42.6 & 80.6 & 2.0 & 27.6& \textbf{83.1} & 8.7\\
    \hline
    \end{tabular}
    \caption{Test BLEU, WinoMT primary-entity accuracy (Acc), and change in second-entity label correspondence $\Delta$L2. We adapt the baseline to a set without tags  (S\&B), or to one of the binary gender-inflection tagging schemes (V1-V4). `Labeled WinoMT' indicates whether WinoMT primary entities are tagged with their reference gender label.  All  results are for rescoring the baseline system gendered-alternative lattices with the listed model.}
    \label{tab:mfresults}
\end{table*}
\begin{table*}
    \centering
    \small  

    \begin{tabular}{|c|cc|cc|cc|cc|cc|cc|}
    \hline
      \textbf{System}   &  \multicolumn{6}{c|}{\textbf{en-de}} &  \multicolumn{6}{c|}{\textbf{en-es}}\\
        &  \multicolumn{2}{c}{Unlabeled} &  \multicolumn{2}{c}{Auto-labeled} & \multicolumn{2}{c|}{Reference labeled}  &  \multicolumn{2}{c}{Unlabeled} & \multicolumn{2}{c}{Auto-labeled} & \multicolumn{2}{c|}{Reference labeled}\\
    \hline

        &  Acc &  $\Delta$L2&    Acc &  $\Delta$L2&   Acc &  $\Delta$L2 &  Acc &  $\Delta$L2  &   Acc &  $\Delta$L2&   Acc &  $\Delta$L2\\
         Baseline   &  60.1  & - &  -& - & - & -& 49.6 & - &-&-  & - & -\\
         S\&B & \textbf{82.3} & 27.4 &- &-& -& -& 66.3 & 29.7 & -& -& -& -\\
         V1 &  81.5 & 26.6& 81.7& 26.5& 81.7 &26.6  & \textbf{67.3}   & 29.6 & 68.5 & 31.2 &69.0 & 26.4\\
         V2 & 71.2 &9.2  & \textbf{83.6} &24.8 & \textbf{84.1}& 24.2 & 52.1 & 3.5& 69.7 & 18.4& 70.9 & 13.2 \\
        V3 &   57.5 & -5.8 &79.9 &\textbf{3.7} &   77.4& \textbf{1.1} &  47.9 & -2.5 & 77.7 & 6.4& 80.6 & \textbf{0.3}  \\
        V4 & 60.5&\textbf{-2.0} & 79.2 & 4.6 & 80.6 & 2.0 &48.5 & \textbf{-0.6} &\textbf{80.6} &12.6 & \textbf{83.1} &8.7 \\
    \hline

    \end{tabular}
    \caption{WinoMT accuracy and  change in second-entity label correspondence for the adaptation schemes in Table \ref{tab:mfresults} when changing how tags are determined for \textbf{WinoMT source sentences}. The primary entity's gender label in each test sentence is either unlabeled, auto-labeled with RoBERTa, or labeled with the reference gender.}
    \label{tab:secondaryresults}
\end{table*}

Table \ref{tab:secondaryresults} lists  accuracy and $\Delta$L2 with and without WinoMT source sentence labeling for the same systems as Table \ref{tab:mfresults}. We also experiment with labeling WinoMT sentences automatically, using RoBERTa to predict the antecedent of the single pronoun in each test sentence -- we note this would not necessarily be as effective in sentences with multiple pronouns.

V1 gives similar performance to S\&B with and without WinoMT labeling.  Removing the possessive pronoun as in V2 decreases accuracy compared to V1 without labeling and slightly increases it with labeling, suggesting removing the source pronoun forces the model to rely on the gender tag. 

Accuracy under V2, V3 and V4 improves dramatically when gender labels are added to WinoMT primary entities. Without labels the accuracies for these systems improve far less or not at all. This is unsurprising, since in these datasets the gender tag is the only way to infer the correct target inflection. Nevertheless some accuracy  improvement is still possible for V2 with neither tags nor possessive pronouns, possibly because the model `sees' more examples of profession constructions in the target language. 

Without test set labels, the V3 and V4 systems have negative $\Delta$L2, implying that the second entity's inflection corresponds to the primary entity label less often than for the baseline. This is not necessarily bad, as they are still low absolute values. Small absolute  $\Delta$L2 indicates that added primary-entity gender signals have little impact on the secondary entity relative to the baseline, which is the desired behaviour. Small negative values are therefore better than large positive values.

Auto-labeling WinoMT source sentences performs only slightly worse than using reference labels. We find that the automatic tags agree with human tags for 84\% of WinoMT sentences, with no difference in performance between masculine- and feminine-labeled sentences, or pro- and anti-stereotypical sentences. This is encouraging, and suggests that the tagged inflection approach may also be applicable to natural text, for which manual labeling is often impractical.

\subsection{Gender-neutral translation}

\begin{table*}
    \centering
    \small  

    \begin{tabular}{|c|c|cc|cc|}
    \hline    
  \textbf{System}  & \textbf{Labeled WinoMT} & \multicolumn{2}{c|}{\textbf{en-de}} &  \multicolumn{2}{c|}{\textbf{en-es}}\\
\hline
         &  & Acc &  $\Delta$L2 & Acc &  $\Delta$L2   \\
         
         Baseline  &$\times$ & 2.7 & - &  4.2	 & -\\ 
         S\&B & $\times$& 13.5& 28.8 &6.4 & 3.9 \\
V1 & \checkmark & \textbf{27.3}& 28.2 &  25.4	& 25.1 \\
V2 & \checkmark & 23.0& 39.6 &  32.1 & 27.5\\
    V3 & \checkmark & 20.2&  18.7 & 38.8 &	 10.0\\
    V4 & \checkmark & 19.4 &  \textbf{4.4} & \textbf{56.5} & \textbf{0.7}\\
    \hline

    \end{tabular}
    \caption{Primary-entity accuracy and second-entity label correspondence $\Delta$L2  on a neutral-label-only WinoMT version. Adaptations sets and lattices are augmented with synthetic neutral articles and nouns. `Labeled WinoMT' indicates whether sentences are tagged with their reference (neutral) gender label.}
    \label{tab:neutral}
\end{table*}
In Table \ref{tab:neutral} we report on systems adapted to the neutral-augmented synthetic sets, evaluated on the neutral-only WinoMT set. We use test labeling for all cases where models are trained with tags -- as with the binary experiments we found that performance was otherwise poor. 

Unsurprisingly, the baseline model is unable to generate the newly defined gender-neutral articles or noun inflections -- the non-zero accuracy is a result of existing WinoMT sentences with neutral entities like `someone'. Adapting on the neutral-augmented S\&B set does little better for en-es, although it gives a larger gain for en-de. This discrepancy may be because the only neutral gender signal in the S\&B source sentences is from the possessive pronoun \emph{their}. In Spanish, which has one gender-neutral third-person singular possessive pronoun, \emph{their} has the same Spanish translation as \emph{his} or \emph{her} and therefore does not constitute a strong gender signal. By contrast in German we add a synthetic singular gender-neutral pronoun, which indicates neutral gender even without tags. This may also explain why V3 and V4 give weaker performance than V1 for German, as these sets no longer contain singular pronouns.

Adding a gender tag significantly improves primary entity accuracy. As with Table \ref{tab:mfresults}, there is little difference in labeled-WinoMT performance when the possessive pronoun is removed. Also as previously, the V3 and V4 `tagged coreference' sets shows far less over-generalization in terms of $\Delta$L2 than the other tagged schemes, although V4 significantly outperforms V3 for en-es on this set. 

We note that primary-entity accuracy is relatively low compared to results for the original WinoMT set, with our best-performing  system reaching 56.5\% accuracy. We consider this unsurprising since the model has never encountered most of the neutral-inflected occupation terms before, even during adaptation, due to the lack of overlap between the adaptation and WinoMT test sets. However, it does suggest that more work remains for introducing novel gender inflections for NMT. 

\section{Conclusions}
Tagging words with target language gender inflection is a powerful way to improve accuracy of translated inflections. This could be applied in cases where the correct grammatical gender to use for a given referent is known, or as monolingual coreference resolution tools improve sufficiently to be used for automatic tagging.  It also has potential application to new inflections defined for gender-neutral language.

However, there is a risk that gender features will be used in an over-general way. Providing a strong gender signal for one entity has the potential to harm users and referents by erasing other entities in the same sentence, unless a model is specifically trained to translate sentences with multiple entities. In particular we find that our V3 system, which is trained on multiple-entity translation examples, allows good performance while minimizing peripheral effects. 

We conclude by emphasising that work on gender coreference in translation requires care to ensure that the effects of interventions are as intended, as well as testing scenarios that capture the full complexity of the problem, if the work is to have an impact on gender bias. 
 \section*{Acknowledgments}
This work was supported by EPSRC grants EP/M508007/1 and EP/N509620/1 and has been performed using resources provided by the Cambridge Tier-2 system operated by the University of Cambridge Research Computing Service\footnote{\url{http://www.hpc.cam.ac.uk}} funded by EPSRC Tier-2 capital grant EP/P020259/1. Work  by R. Sallis during a research placement was funded by the Humanities and Social Change International Foundation.

\bibliographystyle{coling}
\bibliography{refs}

\end{document}